\documentclass[10pt,twocolumn,letterpaper]{article}

\usepackage{iccv}              

\definecolor{iccvblue}{rgb}{0.21,0.49,0.74}
\usepackage[pagebackref,breaklinks,colorlinks,allcolors=iccvblue]{hyperref}
\usepackage{multirow}

\def\confName{ICCV}
\def\confYear{2025}

\title{Instruction-based Image Editing with Planning, Reasoning, and Generation}

\author{Liya Ji\\
HKUST\\
{\tt\small lji@connect.ust.hk}
\and
Chenyang Qi\\
HKUST \\
{\tt\small cqiaa@connect.ust.hk}
\and
Qifeng Chen\footnotemark[1]\\
HKUST \\
{\tt\small cqf@ust.hk}
}

\begin{document}

\twocolumn[{
\maketitle
\begin{center}
    \captionsetup{type=figure}
    \vspace{-2em}
    \includegraphics[width=.98\textwidth]{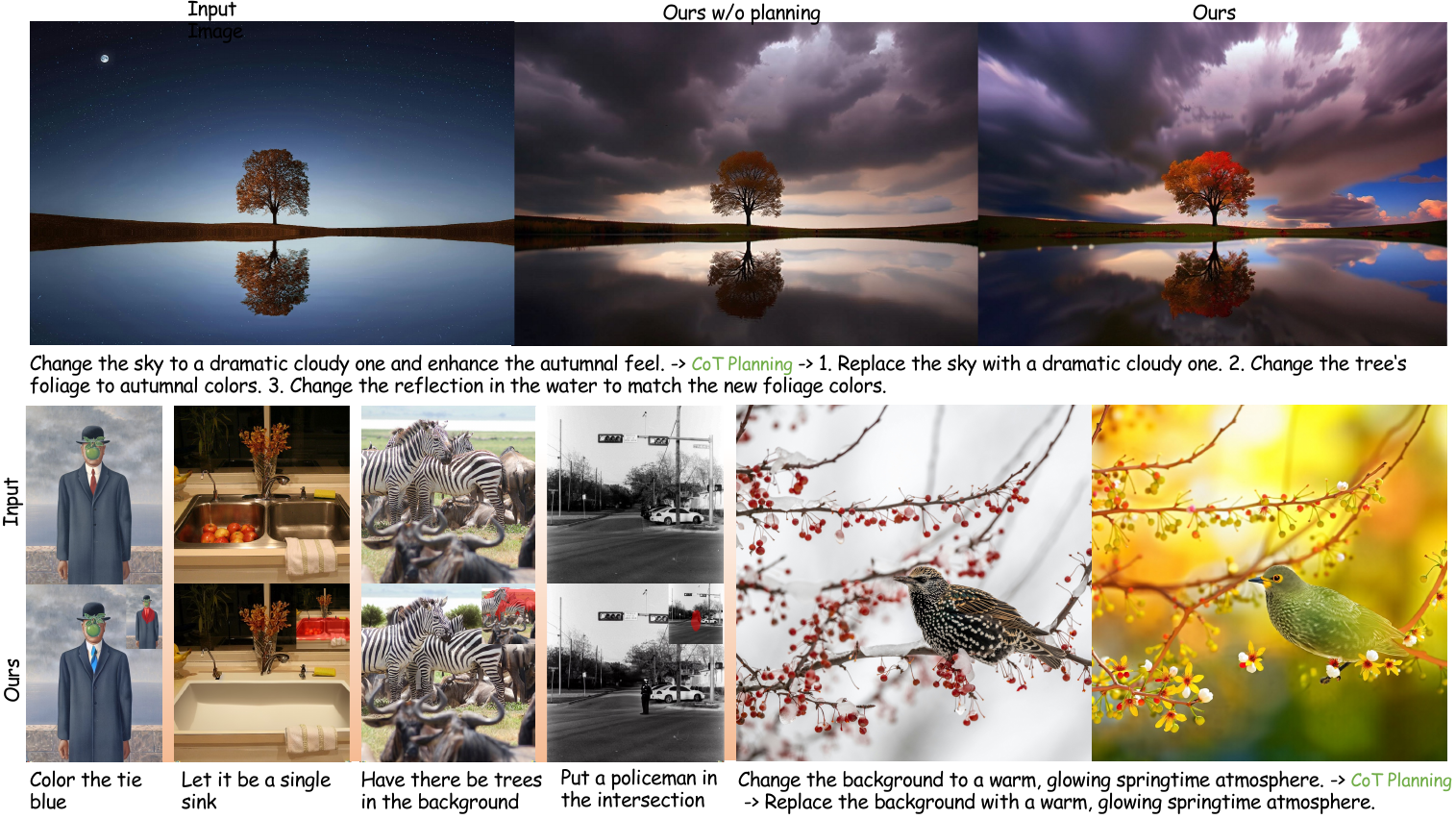}
    \vspace{-1em}
    \captionof{figure}{ 
    We propose an instruction-based editing method with a Planning, Reasoning, and Generation framework that can edit the image with human language, empowered by the (multi-modal) large language model. Row 1 and Row 2 right: Our model could generate more fulfilling contents using instructions obtained by chain-of-thought; Row 2 left: Ours can further reason for the accurate editing region (shown at top right of sub-figures) based on the provided instructions.
    }
\end{center}
}]
\footnotetext[1]{Corresponding authors}
\begin{abstract}

Editing images via instruction provides a natural way to generate interactive content, but it is a big challenge due to the higher requirement of scene understanding and generation.
Prior work utilizes a chain of large language models, object segmentation models, and editing models for this task. However, the understanding models provide only \textit{single} modality ability, restricting the editing quality.
We aim to bridge understanding and generation via a new \textit{multi-modality} model that provides the intelligent abilities to instruction-based image editing models for more complex cases.
To achieve this goal, we individually separate the instruction editing task with the multi-modality chain of thought prompts, \ie, Chain-of-Thought (CoT) planning, editing region reasoning, and editing. 
For Chain-of-Thought planning, the large language model could reason the appropriate sub-prompts considering the instruction provided and the ability of the editing network.
For editing region reasoning, we train an instruction-based editing region generation network with a multi-modal large language model. Finally, a hint-guided instruction-based editing network is proposed for editing image generations based on the sizeable text-to-image diffusion model to accept the hints for generation. Extensive experiments demonstrate our method has competitive editing abilities on complex real-world images. The source code will be publicly available.
\end{abstract}
\section{Introduction}
\label{sec:intro}

Humans are familiar with guiding how to perform a task via instructions since instructions effectively encompass actions and the object that needs to be modified. 
Unlike other settings of language-guided image editing, such as text labels or descriptions of target images, image editing via instructions~\cite{brooks2023instructpix2pix} allows a more user-friendly interaction with concise and accurate action guidance.
This interactive approach expands our imaginations to a world where humans use their language naturally to easily change multimedia resources or artificially generated content.
In addition, instruction-based image editing can be extended with voice control in human-computer interaction scenarios, enhancing the user experience in commercial products.

Previous methods~\cite{brooks2023instructpix2pix,Zhang2023MagicBrush,hui2024hq} tune the text-to-image diffusion model for instruction-based editing in an end-to-end fusion. Several works~\cite{huang2023smartedit, fu2023guiding, yang2024mastering} increase the editing ability of diffusion models~\cite{brooks2023instructpix2pix} with the help of Multi-modality Large Language Models (LLMs), where they~\cite{huang2023smartedit, fu2023guiding} replace the text embedding with MLLMs directly.
These methods have two drawbacks.
Firstly, it increases the workload and requirements of the generation network without requiring additional human prior knowledge, such as splitting a complex problem into several simple tasks.
Secondly, the whole framework is less interpretable 
Then, the editing hints, such as sub-prompts or the editing regions, can also be modified by users easily.

In the real world, image editing via instructions challenges us with higher understanding and reasoning ability requirements due to its complexity.
Some instructions contain abstract concepts, such as ``dramatic'' or ``beautiful'', which can not be understood well by the text encoder only. Besides, the longer instructions contain multiple actions together. Thus, inspired by Chain-of-Thought~\cite{wei2022chain}, we utilize the power of LLM to create a detailed and interoperable prompt to enhance generation ability. Unlike the original Chain-of-Thought, which only contains the text prompt, we are focusing on the image editing tasks; we thus consider the multi-modality prompts, which include the prompt planning, the editing region generation, and the prompt for instruction-based editing. These multi-modal thought chains provide detailed and explainable intermediate results to divide the complex editing task into multi-run editing tasks.

Based on previous motivation, we propose a novel framework, Multimodal Chain-of-Thought Editing, consisting of an MLLM CoT Planner that generates multimodality hints for editing and a hint-guided editing network that produces the final editing results.
The multimodality hints contain the specific sub-prompts and the corresponding editing regions.
We use DeepSeek~\cite{deepseekai2025deepseekr1incentivizingreasoningcapability} Reasoning Model with proper promptings, such as "Let us think step by step," to trigger the chain of thought prompts. 
We aim to instantiate abstract concepts, understand the concept in the context of specific situations, or split complex tasks into simple sub-tasks.
In detail, we not only use the standard way, ``Let us think step by step,'' but also consider the ability to edit the network in the prompt and enable the planner to double-check the answers.
Secondly, inspired by LISA~\cite{lai2023lisa}, we would like the Multi-modal Language Model to reason the edited region directly, given the input image and the editing instructions, leading to a more stable and fine-grained editing quality.
The edited mask, reasoned by the Multimodal Language Model, is an excellent external resource for controlling the generated results spatially.
In addition, we propose a simple but effective hints-guided network by spatially adding the latent spaces of foreground and background images to the noised states in each step of the denoising process.
We found out that the foreground and background images, as the conditions of the diffusion models, could bring effective hint control to the generated results.
We also extend the framework to support classifier-free guidance on three conditions, which, according to the experiments, lead to a slight improvement.

We conduct extensive experiments and achieve state-of-the-art performance on the MagicBrush~\cite{Zhang2023MagicBrush} dataset and the HQEdit-Abstract dataset with abstract concepts instructions, extracted from HQEdit~\cite{hui2024hq}. We also apply our models to real-world cases in the open domain.
Our contributions can be summarized as:
\begin{itemize}
 \renewcommand{\labelitemi}{\textbullet}

     \item We propose a novel framework, Multimodal Chain-of-Thought Editing, consisting of an MLLM CoT Planner that generates multimodality hints for editing and a hint-guided editing network that produces the final results.
    \item We propose an effective hint-guided editing framework by adding the foreground and background images as the conditions of the generation models. 
    \item We create an instruction-based image editing CoT dataset based on MagicBrush. 
    We also conduct extensive experiments on the MagicBrush dataset and the HQEdit-Abstract dataset with state-of-the-art performance and apply our method to the real-world open-domain cases.
\end{itemize}

\section{Related Work}

\begin{figure*}[t!]
    \centering
    \includegraphics[width=\textwidth]{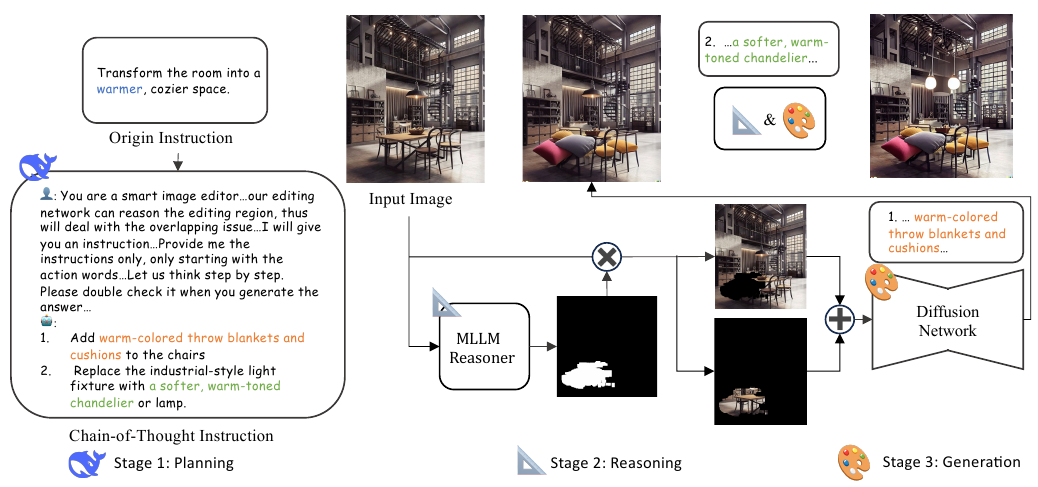}
    \vspace{-2em}
    \caption{Our Multi-modal Chain-of-Thought Editing framework executes image editing through three iterative stages, including planning, reasoning, and generation.
    In stage 1, a \textbf{C}hain-of-\textbf{T}hought Planner decomposes the user prompts to chain-structured refined editing sub-instructions;
    For each sub-instruction, an \textbf{M}LLM  localizes target editing regions (stage 2) via cross-modal reasoning; Then, the conditional Diffusion model \textbf{E}dits the latest image (stage 3) while preserving non-target areas. 
    The system cyclically refines outputs through location reasoning by MLLM and image generation by the Diffusion model until the original plan in stage 1 is completed.
    }
    \vspace{-3mm}
    \label{fig: mccg}
\end{figure*}

\subsection{Instruction-based Image Editing}
Instruction-based image editing~\cite{brooks2023instructpix2pix, 
yildirim2023inst, wang2023instructedit, fu2023guiding, nguyen2023visual, chakrabarty2023learning, mirzaei2023watch, zhang2023hive, huang2023smartedit, guo2023focus, Zhang2023MagicBrush, Geng23instructdiff} provide a more straightforward way for human-like image editing.
InstructPix2Pix~\cite{brooks2023instructpix2pix} is the first work to propose this setting and generate a large instruction-based dataset using the Prompt-to-prompt~\cite{prompt-to-prompt} techniques to control the consistency of the spatial structure.
MagicBrush~\cite{Zhang2023MagicBrush} proposed the instruction-based fine-grained image editing dataset, and they fine-tuned InstructPix2Pix on their dataset, leading to an improvement in editing quality.
Firstly, it is helpful to utilize the masks reasoned by the model as additional information for instruction-based image editing.
Chakrabarty~\etal~\cite{chakrabarty2023learning} tries to use ChatGPT~\cite{gpt4} and GroundingDINO~\cite{liu2023grounding} to generate the mask to filter out a higher-quality dataset on InstructPix2Pix~\cite{brooks2023instructpix2pix}.
InstructEdit~\cite{wang2023instructedit} utilizes the mask provided by chaining ChatGPT with an object-level segmentation model to obtain the editing results.
Qin~\etal~\cite{guo2023focus} extracts the mask from the cross-attention map and provides an attention-guided editing framework.
All of the works~\cite{chakrabarty2023learning, wang2023instructedit, guo2023focus} fail in the scenarios in which we need the editing regions instead of object-level segmentation.
In addition, there are several works~\cite{fu2023guiding, huang2023smartedit} that take advantage of the strong ability of M-LLMs to solve the open-domain challenges in instruction-based image editing.
To the best of our knowledge, we are the first to utilize the Multi-Modality Large Language network to reason the edited regions as a bridge of understanding and generation, and thus relieve the workload of diffusion-based editing models in the instruction-based image editing problem.

\subsection{Multi-Modality LLMs for Vision Tasks}
Multi-modality LLMs are first inspired by GPT-4~\cite{gpt4}, which accepts the image input to the large language models and produces corresponding text output. Based on this intuition, several methods have been proposed for understanding the scenes through neural languages~\cite{liu2023llava,liu2023improvedllava, gong2023multimodalgpt, minigpt}. Besides language tasks, research also involves the computer vision understanding tasks via M-LLM, including the grounding information generation~\cite{xie2023visorgpt}, semantic generation~\cite{seem,lai2023lisa}, and planning~\cite {gao2023assistgpt}.  
Our task is more related to the MLLM-based semantic mask generation. In contrast, we aim to train a network specifically for editing region generation. Then, we can use this hint for the proposed instruction-based image editing network.
Recently, several editing works have been proposed by utilizing M-LLMs~\cite{huang2023smartedit, yang2024mastering, ge2024seed}, which are different from ours.

\subsection{Controllable Generation in Diffusion Models}
Recent works mainly utilize the priors of the diffusion model for conditional generation. ControlNet~\cite{controlnet} and T2I-Adapter~\cite{t2i-adapter} learn to add additional control signal abilities~(\eg, the human pose, the canny edge, and depth) to the stable diffusion. Despite the spatial control, video personalization is another interesting domain of controllable generation. Where we can learn to control the identity of the objects or humans~\cite {celebbasis} via the plugins, such as Dreambooth~\cite{dreambooth}, Custom Diffusion~\cite{customdiffusion}, and IP-Adapter~\cite{ip-adapter}; however, they only work on the pre-trained text-to-image models, which is different from our task, which needs to add controls to the instruction-based editing models.

\section{Method}
We aim to perform general editing on images following complex natural language instructions. 
Given an image $x_0$ and an editing instruction $p$, our model generates the edited image $y$.
Different from previous globally single-stage end-to-end frameworks~\cite {brooks2023instructpix2pix} or directly fine-tune~\cite{Zhang2023MagicBrush, hui2024hq},
we introduce the MLLM Chain-of-Thought planning framework as the bridge of understanding and generation so that we can utilize the powerful reasoning ability of Multi-modality LLMs~\cite{yang2024mastering, zhang2023multimodal}.

\subsection{Image Editing with Multi-modal Chain-of-Thought Prompts}
\label{sec: mccg}
As shown in Fig.~\ref{fig: mccg}, we propose a novel framework for general image editing with MLLM CoT and Conditional Editing, where the MLLM CoT Planner contains multi-modality LLMs that can parse the given instruction and the reference image to produce several sub-prompts. These sub-prompts give detailed thought chain knowledge, including the text and editing mask, concerning the original image and the given prompt. Then, the framework utilizes these multi-domain prompts for generation.

Specifically, suppose we have a CoT planner, ${\mathcal{C}_p(\cdot)}$ for generating sub-prompts and a MLLM reasoner ${\mathcal{C}_m(\cdot)}$ for reasoning the editing regions, these editing hints can be obtained by:
\begin{align}
    C_h = & \{ (\mathcal{C}_m(x_0, p_i), p_i): p_i \in \mathcal{C}_p(p_{x_0}, p, K)\} \nonumber \\
    = & \{(m_i, p_i)\}_{1 \leq i \leq k \leq K},
\end{align}
where $p_{x_0}$ is the global description of the input image $x_0$ and $m_i = \mathcal{C}_m(x_0, p_i)$. 
We denote $k$ as the number of sub-prompts decided by ${\mathcal{C}_p(\cdot)}$ and $K$ as the pre-defined threshold.

After generating the multi-modality prompts from the given image, the conditional generation module is used for instruction-based editing. 
Suppose we have a generative model $\mathcal{G}(\cdot)$ conditioning on reasoned hints $C_h$ and the input image $x_0$.
The edited results $y_0$ is obtained iteratively:
\begin{equation}
    y_{i+1} = \mathcal{G}(y_i, m_i, p_i); \ for\ 1 \leq i \leq k,
\end{equation}
where the edited image $y = y_{k+1}$ and the starting input $y_0 = x_0$.
If we assume that the image quality of $y_{i+1}$ is not lower than that of $y_{i}$ after the operation of conditional generation, the edited quality will remain the same with the input image $x_0$ after several iterations.
However, the assumption is not valid due to the limitations of current state-of-the-art generation models.
Therefore, we set an appropriate small value for $K$ to limit the number of sub-prompts practically.

In detail, we use DeepSeek Reasoning Model~\cite{deepseekai2025deepseekr1incentivizingreasoningcapability} as $C_p(\cdot)$ with proper prompting to trigger the Chain-of-Thought ability.
The prompting details can be found in Figure~\ref{fig: mccg}.
We find that providing the planner with the editing network ability as a prior could remove some unnecessary instructions.
For example, since our editing network can reason about the editing region. Providing this information in the prompt could avoid some position adjustment instructions.
At the same time, letting the planner double-check the answer via adding ``Please double check it when you generate the answer,'' can make the prompt more accurate and stable, especially for dealing with numbering cases.
In addition, we train the M-LLM $C_m(\cdot)$ for reasoning the editing regions(Sec .~\ref {sec: cot}). 
We also perform the learning-based editing $\mathcal{G}(\cdot)$ given the condition set $(y_i, m_i, p_i)$ via a tuned diffusion model~(Sec.~\ref{sec:guided-gen}).

\subsection{Editing Region Reasoning}
\label{sec: cot}
\begin{figure*}[t!]
    \centering
    \includegraphics[width=.9\textwidth]{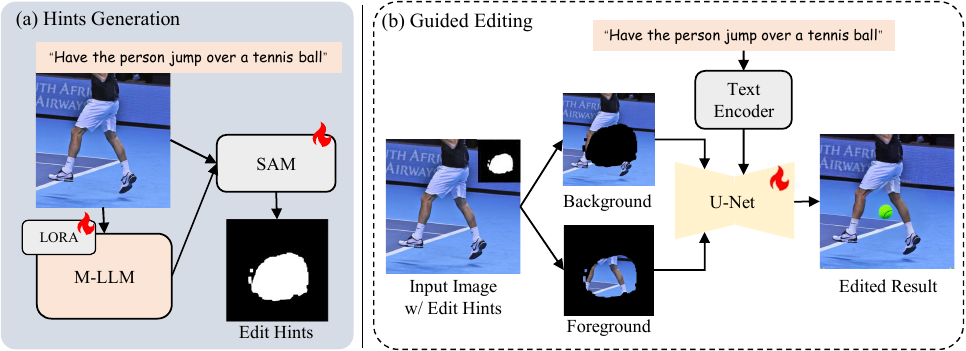}
    \caption{
    (a) We trained a Multi-modality LLM that generates an editing region and enables better localization given the input image and sub-prompt. 
    (b) Given the editing region and sub-prompt reasoned by M-LLM, we further train a conditional generative diffusion model to edit the image with better locality.
    }
    \vspace{-3mm}
    \label{fig:reasonhint}
\end{figure*}
The editing region is a specific kind of mask with high correlations between the input image and the editing instructions according to human opinion.
We argue this region is different from the object-level segmentation and might be a detailed illustration than the object level for specific objects or a meaningless region to put something on. 
Thus, the current universal reasoning segmentation model, \ie, LISA~\cite{lai2023lisa} and SEEM~\cite{seem}, might not work well in these cases.
As shown in Figure~\ref{fig:reasonhint}, if we want to edit an image with the instruction of ``Have the person jump over a tennis ball.'' the segmented region is an area below the legs of the person, instead of the person itself.
In addition, object-level segmentation needs to segment the object precisely, while we only aim to segment an approximate editing region, given an input image, to accept more possibilities.
Since this edited region is not straightforward, it requires the model to be better able to reason about the input instruction. 
Therefore, we need a stronger model that can recognize the image, accept natural language as a prompt, and, most importantly, have the reasoning ability to generate a correct editing area $m_i$. 

Thus, inspired by the recent advantages of multi-modal language model-based image segmentation~\cite{lai2023lisa, seem}, we repurpose the reasoning image segmentation network for our editing region generation task.
In detail, we fix the parameters of the original Multi-Modal LLM~\cite{liu2023llava} and train a LoRA~\cite{lora} to generate the reasoning tokens for segmentation. 
Then, a pre-trained segmentation anything model~(SAM~\cite{kirillov2023segment}) is used to extract the visual feature and generate the reasoning mask with the help of the LLM's output tokens. In this stage, we only train the parameters in LoRA~\cite{lora} and the decoder of SAM~\cite{lai2023lisa} inspired by LISA~\cite{lai2023lisa}. 
We utilize the standard BCE loss to train the network to predict the edit region on the training dataset of MagicBrush~\cite{Zhang2023MagicBrush}. 
After training, this network can be used to infer the editing region from the image and the instruction.


\subsection{Hint-guided Editing Network}
\label{sec:guided-gen}

After getting the condition set $(y_i, m_i, p_i)$,
we propose an efficient network structure to perform the hint-guided image editing. 
In detail, we 
utilize the structure of Stable Diffusion~\cite{stable-diffusion} as the network structure, where a denoising U-Net $\epsilon_\theta(\cdot)$ is trained for image editing via the paired dataset using Prompt-to-Prompt~\cite{prompt-to-prompt}. The CLIP text encoder $\varepsilon(\cdot)$ is frozen to accept the instruction-level guidance. 
Aiming to get better control based on the editing hints, we first compute the foreground image $x_f$ and background image $x_b$ with the edit region $m_i$ by:
\begin{align}
    x_f = &y_i \odot m_i, \\
    x_b = &y_i \odot (1 - m_i).
\end{align}
Then we concatenate the foreground image and background image as an additional spatial condition as the input of Diffusion U-Net $\epsilon_\theta(\cdot)$.
We also encode $x_f$ and $x_b$ into the latent space using the latent encoder $\xi(\cdot)$ before feeding into the denoising network.
We modify the standard diffusion loss~\cite{stable-diffusion} to optimize our network:
\begin{equation}
\small
\begin{split}
     E_{\xi(y_i), \varepsilon(p_i), \xi(x_f), \xi(x_b), t, \epsilon \sim \mathcal{N}(\mathbf{0}, \mathbf{I})}[||\epsilon - \epsilon_{\theta}(z_t, t, \varepsilon(p_i), \xi(x_f), \xi(x_b)||^2_2],
\end{split}
\end{equation}
So we only need to modify the weights of the first convolution layer to fit this difference. In the training process, we use the ground truth mask as the input since it can provide more reliable guidance. In testing, we perform the editing based on the reasoning results from the hints generation network.

\subsection{Classifier-free Guidance for Three Conditions}
\label{method: cfg}
The denoising diffusion model with conditions will decrease the diversity of the generated results. 
Adding classifier-free guidance aims to let the model maintain the original generation ability by randomly dropping some conditions during the training process.
It is even more important with the increase in the number of conditions.
In our hints-guided network, the denoising network $\epsilon_\theta(z_t, x_f, x_b, p_i)$ has three conditions, which denote the foreground image $x_f$, the background image $x_b$, and the instruction $p_i$ separately.
We extend two conditions~\cite{brooks2023instructpix2pix} to three conditions as below:
\begin{equation}
\label{eq: cfg}
\begin{split}
    \epsilon_\theta(& z_t, x_b, p_i, x_f) = \epsilon_\theta(z_t, \phi, \phi, \phi) \\
                                &  + s_f (\epsilon_\theta(z_t, x_f, \phi, \phi) - \epsilon_\theta(z_t, \phi, \phi, \phi)) \\
                                &  + s_p (\epsilon_\theta(z_t, x_f, x_b, \phi) - \epsilon_\theta(z_t, x_f, \phi, \phi))  \\
                                &  + s_b (\epsilon_\theta(z_t, x_f, x_b, p_i) - \epsilon_\theta(z_t, x_f, x_b, \phi)),
\end{split}
\end{equation}
where $s_f, s_b, s_p$ denote the guidance scales for foreground image condition, background image condition, and text condition, respectively.
From Equation~\ref{eq: cfg}, we can see four situations. During the training process, we randomly drop the instruction condition at $5\%$, drop both the background image and instruction at $5\%$, and drop all three conditions at $5\%$. 
\begin{figure*}[t!]
    \centering
    \includegraphics[width=.95\textwidth]{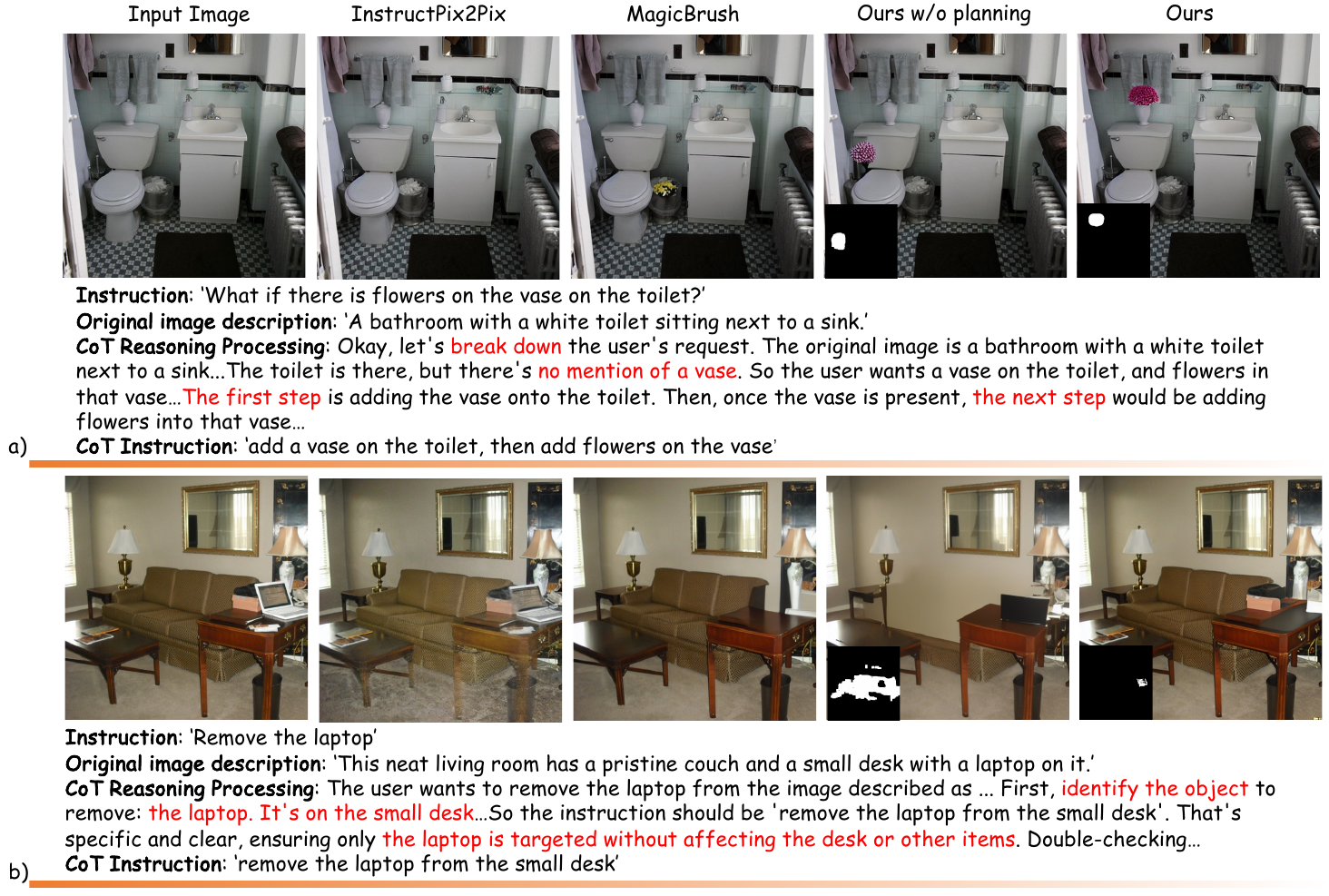}
    \vspace{-1em}
    \caption{\textbf{Examples of our method of the instruction-based image editing on MagicBrush~\cite{Zhang2023MagicBrush}}. Editing regions reasoned by M-LLM are shown in the bottom left corner of our editing results.
    Under the examples, we show Chain-of-thought (CoT) planning, which helps to understand some concepts or break down the tasks.
    }
    \vspace{-1em}
    \label{fig:examples}
\end{figure*}
\section{Experiments}
\subsection{Datasets and Pretrained Models}
We train our methods on MagicBrush~\cite{Zhang2023MagicBrush}, an instruction-based dataset with high quality for a local image editing dataset, which also provides the edited mask compared to other instruction-based datasets.
We trained our MLLM reasoner and hints-guided editing network on the released train dataset containing 4,600 input images.
In addition, although this dataset is high-quality, the number of instances is limited due to the high labor cost.
We augment this dataset five times at last, as introduced in our method, and extend the training dataset to contain 78,000 input image editing pairs, a significant number for training an editing model.
The details of augmenting data can be found in the Appendix.

We evaluate the performance of our model on two datasets.
One is the released test dataset of MagicBrush.
In addition, we have built a small dataset with 100 samples, containing abstract concepts, such as ``warm'', ``dramatic'', and ``playful'', extracted from HQEdit~\cite{hui2024hq}.
We evaluate the effectiveness of our whole framework, denoted as ``ours,'', consisting of planning, reasoning, and generation on the HQEdit-abstract dataset.
In addition, for the description of the input image in the CoT planning phrase, we use the global description provided by the MagicBrush dataset and utilize gpt-4o~\cite{gpt4} to generate the descriptions for HQEdit~\cite{hui2024hq}
or other open-domain cases.

We utilize three pretrained models and fine-tune our method on them.
For the hints-generation network, we utilize SAM~\cite{kirillov2023segment} for segmentation and LLava-7b~\cite{liu2023llava, liu2023improvedllava} for multi-modal LLMs.
For the hints-guided network, we utilize instructPix2Pix~\cite{brooks2023instructpix2pix} as the initialization for the weights of the denoising U-Net parts.


\begin{table*}[h]
\begin{minipage}[t]{0.6\textwidth}
\makeatletter\def\@captype{table}
\setlength{\tabcolsep}{3pt}
\renewcommand{\arraystretch}{1.27}
  \caption{\textbf{Quantitative results of our instruction-based editing model on MagicBrush~\cite{Zhang2023MagicBrush} test datasets}. We calculate the CLIP~\cite{radford2021learning} similarity using global and local descriptions separately. The total score is the average score among all metrics.}
\label{tab: main-table}
\begin{tabular}{l | c | c c c c}
  \toprule[1pt]
    Methods & {Total} & {CLIP-I$\uparrow$} & {DINO-I$\uparrow$} & {CLIP-T$\uparrow$ } & {CLIP-T $\uparrow$} \\
     & Score &   &  &  (Global)  & (Local) \\
    \midrule
    InstructPix2Pix~\cite{brooks2023instructpix2pix} & 0.5457 & 0.8595 & 0.7501 & 0.2942 & 0.2791 \\
    InstructDiffusion~\cite{Geng23instructdiff} & 0.5754 & 0.8980 & 0.8226 & 0.2997 & 0.2814 \\
    MagicBrush~\cite{Zhang2023MagicBrush} & 0.5853 & 0.9080 & 0.8443 & \textbf{0.3035} & \textbf{0.2855} \\
    HIVE~\cite{zhang2023hive} & 0.5493 & 0.8599 & 0.7681 & 0.2928 & 0.2762 \\
    Ours w/o planning & 0.5881 & 0.9117 & 0.8554 & 0.3026 & 0.2826 \\
    Ours & \textbf{0.5904} & \textbf{0.9172} & \textbf{0.8658} & 0.2995 & 0.2789 \\
    \bottomrule[1pt]
  \end{tabular}
\end{minipage}
\hfill
\begin{minipage}[t]{0.35\textwidth}
\makeatletter\def\@captype{table}
\renewcommand{\arraystretch}{0.975}
\setlength{\tabcolsep}{1.5pt}
   \caption{\textbf{Quantitative results of our multimodal Chain-of-Thought editing framework on HQEdit-Abstract Dataset.} We show the voting ratio regarding the correctness of editing quality and the subject rating of abstract concepts.}
   \begin{tabular}{l c c}
  \toprule[1pt]
    Methods & {Editing $\uparrow$} & {Abstract$\uparrow$}  \\
     & Quality & Score \\
    \midrule
    MagicBrush~\cite{Zhang2023MagicBrush} &20.71\% & 22.76\%  \\
    HQEdit~\cite{hui2024hq} & 23.53\% & 23.01\% \\
    Ours w/o planning & \textbf{28.64}\% & 24.80\%\\
    Ours & 27.10\% & \textbf{29.41\%}  \\
    \bottomrule[1pt]
    \end{tabular}
    \label{tab: hqedit}
\end{minipage}
\end{table*}
\subsection{Baselines and Metrics}
We have picked InstructPix2Pix~\cite{brooks2023instructpix2pix}, InstructDiffusion~\cite{Geng23instructdiff}, MagicBrush~\cite{Zhang2023MagicBrush}, and HIVE~\cite{zhang2023hive}, HQEdit~\cite{hui2024hq} as our baselines. 
We run the checkpoints provided publicly on MagicBrush's dev dataset, keeping the hyperparameters the same as those in the released codes.
Those works do not need users to provide a mask at the test time.

Following previous image editing methods~\cite{Zhang2023MagicBrush}, we use the embeddings of the CLIP~\cite{radford2021learning} and DINO~\cite{caron2021emerging} to calculate the cosine similarity between the generated output and the ground-truth output provided by the dataset, which are denoted as CLIP-I and DINO-I, respectively.
In addition, we utilize the global and local descriptions of the ground-truth output. We also use CLIP to calculate the similarity between the generated output and the description, which are denoted as CLIP-T~(Global) and CLIP-T~(Local).
As for the HQEdit-abstract dataset, we conduct a user study of around 23 workers.
Each worker needs to answer two questions for each pair. One is for the correctness of editing quality, and another one is for the subject rating about the consistency with the corresponding abstract concepts, denoted as abstract-score.

\subsection{Implementation details}
We train each model individually.
For the hints-guided network, we trained our model from the pretrained InstructPix2Pix~\cite {brooks2023instructpix2pix} at the resolution 256 $\times$ 256 with epoch 200 on the training set of the MagicBrush~\cite{Zhang2023MagicBrush}. The batch size is 2 for 8 V100 GPUs. We set the learning rate to 1$e^{-4}$. We use the SD-XL~\cite{podell2023sdxl} as the generative fill model and set the probability threshold $\gamma$ to pick the augmented images as 50\%.
We use the ground-truth mask for training and the mask predicted by the hints-generation network for inference.
The DDIM steps for inference are set to 100 as the original instructPix2pix~\cite{brooks2023instructpix2pix}. 
For the hints-generation network, the total training step is 2500 with a learning rate of 1$e^{-4}$ with the ground truth mask from MagicBrush~\cite{Zhang2023MagicBrush}.
The threshold K in CoT planning is set to 3.

\subsection{Experimental Results}
\subsubsection{Experimental Results on MagicBrush}
Table~\ref{tab: main-table} shows the quantitative results of our methods, where the proposed method shows the state-of-the-art performance compared with baselines. We also give some visual results in Figure~\ref{fig:examples} to prove the efficiency of the proposed method. Since we infer the edit hints from a multi-modality LLM, the mask gives accurate hints of the edited region, which performs better than previous methods.

Based on the proposed editing region-based fine-tuning, the proposed method can accurately reason the location. 
Another example is that the original InstructPix2pix~\cite{brooks2023instructpix2pix} and MagicBrush~\cite{Zhang2023MagicBrush} can not perform local editing well since they do not have an explicit mask.
In addition, for some complex examples, instead of changing too much, some baselines, lacking powerful editing ability, choose not to change the input image, leading to a low CLIP-T(Local) score.
In conclusion, our method could edit the image at an appropriate level with the help of reasoning hints.

\begin{figure*}[t!]
    \centering
    \includegraphics[width=\textwidth]{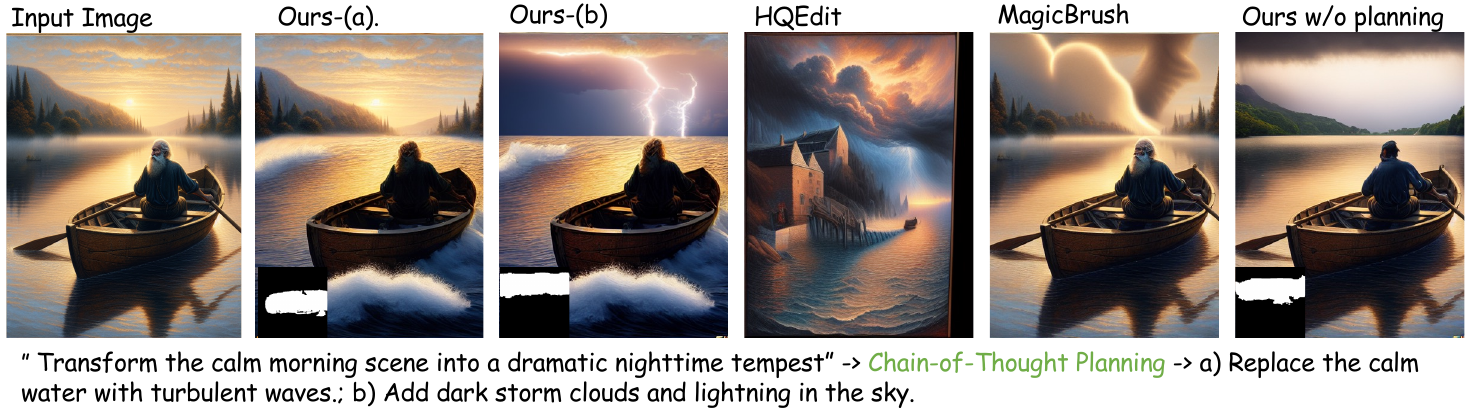}
    \vspace{-2em}
    \caption{\textbf{Examples of our Multimodal Chain-of-Thought Editing Framework on HQ-Abstract.}
    The abstract topic is `dramatic''.
    Our CoT planning with multimodal LLMs could instantiate the abstract instruction into more specific details. The editing area is shown in the bottom left of each image.
    }
    \label{fig: cot_example}
\end{figure*}
\subsubsection{Experimental Results on HQEdit-Abstract}
Table~\ref{tab: hqedit} shows the user study results of our methods, compared with HQEdit~\cite{hui2024hq}, MagicBrush~\cite{Zhang2023MagicBrush}, and our method without prompt planning.
The table shows that our method produces better results than the framework without the help of M-LLMs.
However, the score of editing quality has slightly decreased due to the decrease in image quality after the operation of conditional generation, illustrated in Sec~\ref{sec: mccg}.
However, due to the powerful knowledge brought by M-LLMs, the edited results with our framework could bring more plentiful results to building an atmosphere regarding abstract concept topics.

Figure~\ref{fig: cot_example} shows two examples of our framework.
The abstract topic is ``warm'' for the first line and ``dramatic'' for the second line.
Our CoT planning with multimodal LLMs could instantiate the abstract instruction into more specific details.
For example, if we want to create a dramatic nighttime tempest, we should first add turbulent waves and then dark storm clouds and lightning.
This information could not be encoded only from the abstract instructions.



\subsection{Ablation Study}
\label{exp: ablation}
\begin{figure}[t]
    \centering
    \includegraphics[width=.95\columnwidth]{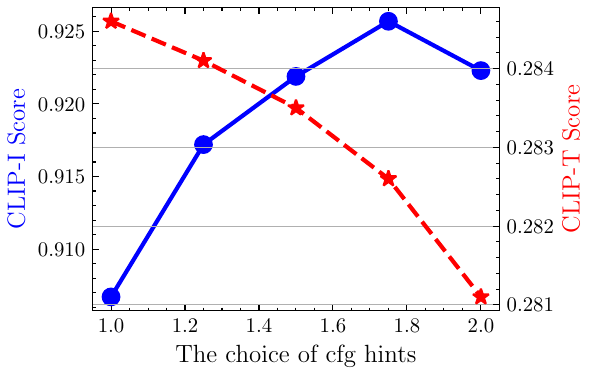}
    \vspace{-1em}
    \caption{\textbf{The influence of editing hints-related CFG.} We evaluate the model performance on MagicBrush dev dataset with the original instruction. We have set the text CFG to a fixed value 7.5.}
    \vspace{-5mm}
    \label{fig:cfg_hint}
\end{figure}

\begin{table}[t]
\centering
\setlength{\tabcolsep}{3pt}
\renewcommand{\arraystretch}{1.2}
\scalebox{0.95}{
  \begin{tabular}{l c c c }
  \toprule[1pt]
    Ablation & {Methods} & {CLIP-I$\uparrow$}  & {CLIP-T$\uparrow$} \\
    \midrule
    \multirow{3}{*}{Hints method} & {Pretrained LISA}& {0.9081} & {0.2766} \\
     & Ground Truth & \textbf{0.9277} &  \textbf{0.2838} \\
     & Ours & {0.9219} & {0.2835}\\
    \midrule
    \multirow{4}{*}{The ratio of}& {0} & \textbf{0.9314} & {0.2771} \\
    &{0.25} & {0.9290} & {0.2796} \\
    {augmented data} &{0.5} & {0.9219} & \textbf{0.2835} \\
    &{0.75} & {0.9206} &{0.2796} \\
    \bottomrule[1pt]
  \end{tabular}
  }
\vspace{-3mm}
\caption{\textbf{Ablation study on the hints generation method and the ratio of the utility of augmented data in the training process.} We evaluate the performance on MagicBrush dev dataset with the original instruction. CLIP-T is calculated via the local description.
Both the hint generation and the augmented dataset benefit the quality of generation results. }
\vspace{-5mm}
\label{tab: ablation_small}
\end{table}
Firstly, we conduct the ablation study on the classifier-free guidance. 
We added extra conditions for classifier-free guidance, and thus, we varied the values of the Classifier-free Guidance~(CFG) foreground image and background image simultaneously from 1.0 to 2.0 with a fixed value for the CFG text of 7.5.
Figure~\ref{fig:cfg_hint} shows the CLIP-I and CLIP-T for local description with different hints-related CFG at the test time.
We can see that more control of the hints-related conditions will increase the CLIP-I scores since we can better maintain the unmasked area of the input images.
However, the larger control of hints-related conditions will also hurt the generation ability of our editing model and the diversity of our edited results with the decrease in CLIP-T scores.

In addition, we conduct the ablation study on the mask-generated model. 
We compare our method with the pretrained LISA model and the ground truth of the mask provided by the MagicBrush dataset.
Finally, we also conducted an ablation study on the random probability $\gamma$, indicating how to use our generated data in our training process. The details can be found in Table~\ref{tab: ablation_small}.

\subsection{Flux Editing Models with CoT Planning}
\begin{figure}
  \centering
    \includegraphics[width=\columnwidth]
    {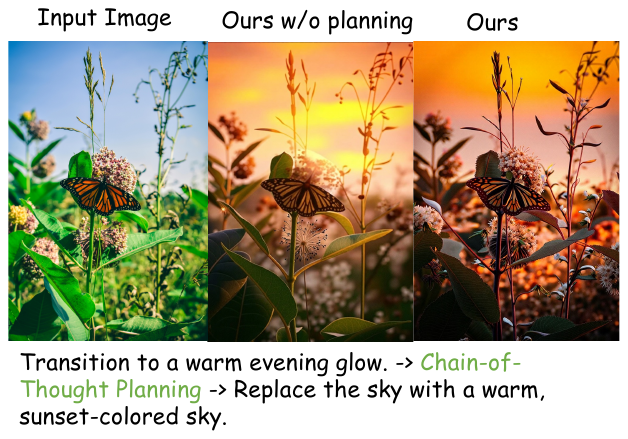}
    \vspace{-7mm}
    \caption{Examples of Flux editing network with Chain-of-Thought planning.}
    \vspace{-7mm}
    \label{fig: fig_flux}
\end{figure}
We extend the Chain-of-Thought Planning on Flux-based Editing Models.
Based on the Flux~\cite{flux2024} text-to-image generation model, we train the Flux editing model with the Controlnet~\cite{controlnet} framework.
Figure~\ref{fig: fig_flux} shows an example of flux editing models with CoT planning.
We find that with the help of CoT planning, the edited results have more alignment with the input image and more fulfilling content, considering the instruction.


\section{Conclusion}
In this paper, we propose a novel framework, called Multimodal Chain-of-Thought Editing, as the bridge between scene understanding and scene editing. 
This framework consists of three parts: a Chain-of-Thought planner, an MLLM reasoner, and a hints-guided editing network.
The experiments show the advantage of the proposed method over the state-of-the-art method on the benchmark of \ie, MagicBrush~\cite{Zhang2023MagicBrush} and a small dataset with abstract instructions extracted from HQEdit~\cite{hui2024hq}. 
In the future, the current MLLM reasoner still faces the issue of inaccurate editing regions. One possible direction is utilizing the current Chain-of-Thought LLMs to improve the reasoning ability.
As for the base generation model, it is promising that extending our whole framework to the Flux model comprehensively.

\section{Acknowledgements}
This project was supported by the Research Grant Council of the Hong Kong Special Administrative Region under grant number 16203122.

{
    \small
    \bibliographystyle{ieeenat_fullname}
    \bibliography{main}
}

\end{document}